\documentclass[a4paper,conference]{IEEEtran}
\usepackage{tabularx,booktabs}
\newcolumntype{Y}{>{\centering\arraybackslash}X}
\newcolumntype{R}{>{\raggedleft\arraybackslash}X}
\usepackage{arydshln}
\usepackage{tabu}
\usepackage{ctable}
\usepackage{enumerate,multicol}
\usepackage{multirow}
\usepackage{multicol}
\usepackage{enumitem}
\usepackage{amsmath,array,graphicx}
\usepackage{gensymb}
\usepackage{float}
\usepackage{caption}
\usepackage{subcaption}
\usepackage{balance}
\usepackage{xspace}
\usepackage{hyperref}
\usepackage{xcolor}
\hypersetup{
    colorlinks,
    linkcolor={red!50!black},
    citecolor={blue!50!black},
    urlcolor={black}
}

\newcommand{\ie}{\textit{i.e.}, }

\begin{document}
\title{Relevance Detection in Cataract Surgery Videos by Spatio-Temporal Action Localization}

\author{\IEEEauthorblockN{Negin Ghamsarian\IEEEauthorrefmark{1},
Mario Taschwer\IEEEauthorrefmark{1}, Doris Putzgruber-Adamitsch\IEEEauthorrefmark{2}, 
Stephanie Sarny\IEEEauthorrefmark{2},  Klaus Schoeffmann\IEEEauthorrefmark{1}}
\IEEEauthorblockA{\IEEEauthorrefmark{1}Department of Information Technology, Alpen-Adria-Universit\"at Klagenfurt\\
\IEEEauthorrefmark{2}Department of Ophthalmology, Klinikum Klagenfurt\\
Email: \IEEEauthorrefmark{1}\{negin, mt, ks\}@itec.aau.at, \IEEEauthorrefmark{2}\{doris.putzgruber-adamitsch, stephanie.sarny\}@kabeg.at}}
\maketitle

\maketitle

\begin{abstract}
In cataract surgery, the operation is performed with the help of a microscope. Since the microscope enables watching real-time surgery by up to two people only, a major part of surgical training is conducted using the recorded videos. To optimize the training procedure with the video content, 
the surgeons require an automatic relevance detection approach. In addition to relevance-based retrieval, these results can be further used for skill assessment and irregularity detection in cataract surgery videos. In this paper, a three-module framework is proposed to detect and classify the relevant phase segments in cataract videos. Taking advantage of an idle frame recognition network, the video is divided into idle and action segments. To boost the performance in relevance detection, the cornea where the relevant surgical actions are conducted is detected in all frames using Mask R-CNN. The spatiotemporally localized segments containing higher-resolution information about the pupil texture and actions, and complementary temporal information from the same phase are fed into the relevance detection module. This module consists of four parallel recurrent CNNs being responsible to detect four relevant phases that have been defined with medical experts. The results will then be integrated to classify the action phases as irrelevant or one of four relevant phases. Experimental results reveal that the proposed approach outperforms static CNNs and different configurations of feature-based and end-to-end recurrent networks.

\end{abstract}
\IEEEpeerreviewmaketitle

\section{Introduction}
\label{sec: Introduction}
Cataract surgery is not only the most frequent ophthalmic surgery but also one of the most common surgeries worldwide~\cite{JFCS}. This surgery is the procedure of returning a clear vision to the eye by removing its cloudy natural lens and implanting an artificial lens. This procedure is conducted with the help of a binocular microscope that provides a three-dimensional magnified and illuminated image of the eye. The microscope contains a  mounted camera, which records and stores the whole surgery for several postoperative objectives. The major objective is to accelerate the teaching process for trainee surgeons, as watching the live surgery is only possible for up to two people. A faster training process results in reducing the complications during and after surgeries and diminishing the surgical risks for less-experienced surgeons. As a concrete example, the risk of developing \textit{reactive corneal edema} after surgery for novice surgeons is reported to be 1.6 times higher than that for experienced expert surgeons~\cite{IOSE2017}.  Accordingly, it is vital to accelerate the teaching and training process for a surgical technique. 

\begin{figure*}[!th]
    \centering
    \includegraphics[width=0.8\textwidth]{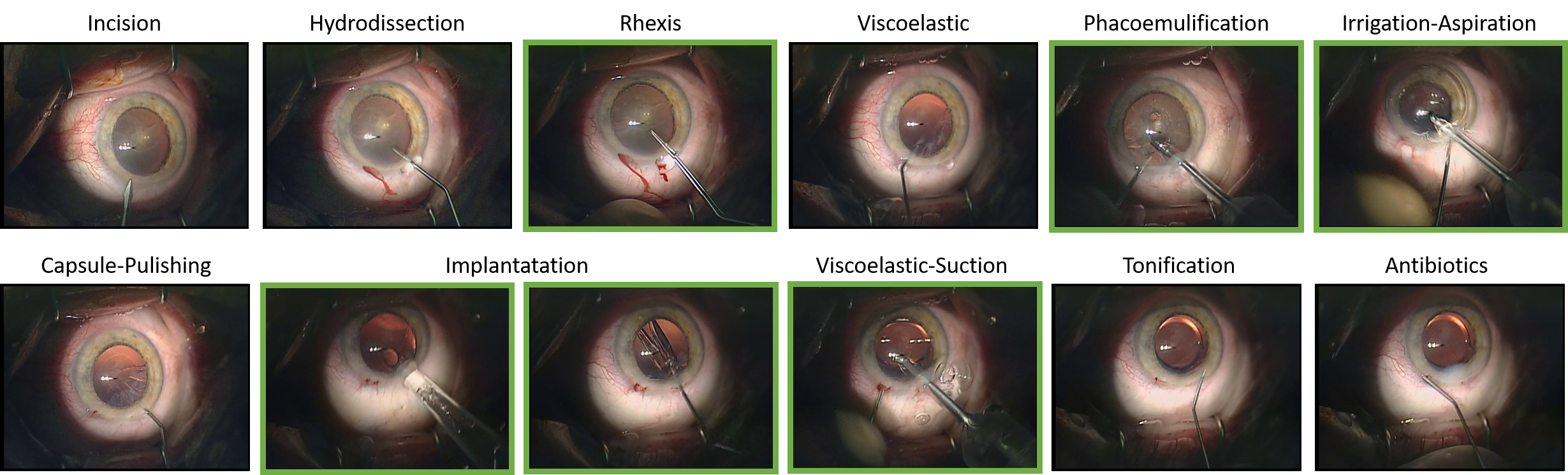}
    \caption{Sample frames of action phases in cataract surgery. Medically relevant phases are illustrated with green borders.}
    \label{fig: samples}
\end{figure*}

To systematically study the relevant surgical phases and investigate the irregularities in cataract surgeries, the sole videos are not sufficient. Dealing with tens of thousands of videos to find particular relevant phases and irregularities is burdensome and time-consuming. Hence, the surgeons require a surgical video exploration system which can shorten the surgical training curve (\ie reducing the training time by enabling fast search) and subsequently result in improved overall surgical outcomes. One of the fundamental components of such a system is an automatic phase segmentation and classification tool~\cite{RBE}. %Accordingly, many studies have sought phase and action recognition in surgical videos. Phase recognition can also pave the way to meet various demands regarding modernizing the operation rooms. 

A cataract surgery video regularly consists of eleven action phases: incision, hydrodissection, rhexis, viscoelastic, phacoemulsification, irrigation-aspiration, capsule polishing, lens implantation, viscoelastic-suction, tonification, and antibiotics. However, not all of the aforementioned phases are equally relevant to clinicians. They consider only \textit{rhexis}, \textit{phacoemulsification}, \textit{irrigation-aspiration} with \textit{viscoelastic-suction}, and \textit{lens implantation} as important from a medical perspective. The intraoperative complications resulting from these phases are reported to have a higher rate compared to that of the irrelevant phases~\cite{ICR2018}. Hence, detecting the aforementioned relevant phases in cataract surgery videos is of prime concern. 

Fig.~\ref{fig: samples} displays sample frames from relevant and irrelevant phases in cataract surgery videos. Designing an approach to detect and classify the relevant phases in these videos with the frame-wise temporal resolution is quite challenging due to several reasons: (i) These videos may contain defocus blur due to manual adjustment of the camera focus~\cite{DCS}. (ii) Unconscious eye movements and fast motion of the instrument lead to motion blur and subsequently dilution of the discriminative and salient spatial segments.
(iii) As shown in Fig.~\ref{fig: samples}, the instruments that are regarded as the major difference between relevant phases are highly similar in some phases. This similarity can result in a narrow inter-class deviation in a trained classifier. (iv) The stored videos do not contain any metadata to be used as side information. 

In this paper, we propose a novel deep-learning-based approach to detect the relevant phases in cataract surgery videos. Our main contributions are listed as follows:
\begin{enumerate}
    \item We present a broad comparison between many different neural network architectures for relevant phase detection in cataract surgery videos, including static CNNs, feature-based CNN-RNNs, and end-to-end CNN-RNNs. This comparison enables confidently scaling up the best approach to various types of surgeries and different datasets. % We have also presented an apple-to-apple comparison between the end-to-end and feature-based (two-step CNN-LSTM) frameworks.
    \item We propose a novel framework for relevance detection in cataract surgery videos using cooperative localized spatio-temporal features.
    %\item 
    \item To enable utilizing complementary temporal information for relevance detection, idle frame recognition is proposed to temporally localize the distinct action phases in cataract surgery videos. Besides, using a state-of-the-art semantic segmentation approach, the cornea region in each frame is extracted to localize the spatial content in each action phase. In this way, we avoid inputting the substantial redundant and misleading information to the network as well as providing higher resolution for the relevant spatial content.
    \item Together with this work, we publish a dataset containing the training and testing videos with their corresponding annotations. This public dataset will allow direct comparison to our results. 
    \item The experimental results confirm the superiority of the proposed approach over static, feature-based recurrent, and end-to-end recurrent CNNs. 
\end{enumerate}

In Section~\ref{sec: relatedwork}, we position our work in the literature by reviewing the state-of-the-art for phase recognition in surgical videos. In Section~\ref{sec: Methodology}, we describe the shortcomings of existing approaches and give a brief explanation of \textit{Mask R-CNN}. We then delineate the proposed relevance detection framework based on action localization termed as \textit{LocalPhase}. The experimental settings are explained in Section~\ref{sec: Experimental Setup} and experimental results are presented in section~\ref{sec: Experimental Results}. The paper is finally concluded in Section~\ref{sec: Conclusion}.     

% LocalPhase was just defined in the aforementioned paragraph. It should not be removed.

\section{Related Work}
\label{sec: relatedwork}

Phase recognition in surgical videos can be categorized into (1) feature-based and (2) deep-learning-based approaches. The approaches corresponding to the first generation extract hand-engineered features to be further used as the inputs to the traditional machine learning methods. Some traditional methods exploit features such as texture information, color, and shape~\cite{FRHS}. The major discriminative features in different phases of surgical videos are the specific tools used in these phases. Accordingly, some methods based on conditional random fields~\cite{RTS} or random forests~\cite{RFPD} have taken advantage of tool presence information for phase recognition. Another method exploits Dynamic Time Warping (DTW), Hidden Markov Models (HMMs), and Conditional Random Fields (CRFs) along with the tool presence signals for surgical video analysis~\cite{RTM}. Some other methods adopt hand-crafted features for bag-of-features (BoF) or multiple kernel learning (MKL)~\cite{SGC}. 

Hand-crafted features may provide sub-optimal classification results or fail to provide robust classification. Automated feature extraction thanks to deep neural networks have proved to provide more optimal and robust classifiers for complicated problems. Hence, deep-learning-based approaches have drawn much attention from the medical imaging community in recent years.  EndoNet~\cite{EndoNet} leverages AlexNet trained for tool recognition as the feature extractor for Hierarchical Hidden Markov Model (HHMM) to classify the phases in laparoscopic surgeries. EndoRCN~\cite{EndoRCN} uses the same technique with ResNet50 as the backbone network and recurrent layer instead of HHMM for phase recognition. DeepPhase~\cite{DPS} exploits tool features as well as tool classification results of ResNet152 as the input for RNNs to perform phase recognition. A comparison between the performance of stacked GRU layers~\cite{GRU} and LSTM~\cite{LSTM} in phase recognition revealed that GRU is more successful in inferring the phase based on binary tool presence. On the other hand, LSTM performs better when trained on tool features. Authors of~\cite{SV-RCNet} proposed an end-to-end CNN-LSTM to exploit the correlated spatio-temporal features for phase recognition in cholecystectomy videos. In another study, the most informative region of each frame in laparoscopic videos is extracted using "Adaptive Whitening Saliency" (AWS) to be used as input to the CNN \cite{SPR}. To compensate for the lack of enough labeled data, various methods for self-supervised pre-training of CNNs are proposed~\cite{SPR}. In these methods, the inherent labels in the dataset are used to train a network based on \textit{(a)} contrastive loss, \textit{(b)} ranking loss, and \textit{(c)} first and second-order contrastive loss.

Regarding action recognition in regular videos, DevNet~\cite{DevNet} has achieved promising results by adopting a spatio-temporal saliency map. LSTA~\cite{LSTA} proposes an attention mechanism to smoothly track the spatially relevant segments in egocentric activities. R(2+1)D~\cite{R(2+1)D} introduces a novel spatiotemporal convolutional block to boost the performance in action recognition. 
%To address the under-performing of 3D CNNs in case of insufficient training examples, a gate-shift module is introduced to turn a static lightweight CNN into a spatiotemporal feature extractor~\cite{GSM}. 
While the aforementioned approaches provide outstanding results, they are specifically designed for action recognition in case of static backgrounds. In contrast, not only do we have a dynamic deformable background, but also the action is not calibrated in the video (\ie the actions are being conducted inside the moving cornea). This makes action recognition in cataract videos more challenging.

\section{Methodology}
\label{sec: Methodology}
\subsection{Shortcomings of Existing Approaches}
Despite using the state-of-the-art baselines and showing good performance in phase recognition, the existing approaches suffer from several flaws, which are discussed as follows: 

\vspace{0.50em}
\noindent \textbf(I) Cataract surgery videos which contain irregularities in the succession of phases are of major importance for clinicians. Hence, it is expected that the trained model can recognize the phases in case of irregularities in the order and duration of them on a frame-level basis. Such a network should not be trained on the time-related and neighboring-phase-related information. Otherwise, the network memorizes the succession of phases and the relative time index of each phase in regular surgeries. Consequently, the network will fail to accurately infer the phases in irregular surgeries~\footnote{As a concrete example, a rarely occurred irregularity in cataract surgery videos is \textit{hard-lens} condition where the surgeons have to perform \textit{phacoemulification} phase in two stages. In the second stage, \textit{phacoemulsification} is performed after \textit{irrigation-aspiration}, while in a regular cataract surgery, the \textit{irrigation-aspiration} phase is always followed by \textit{capsule-polishing} and \textit{viscuelastic}.}. 

\vspace{0.50em}
\noindent \textbf(II) Previous methods in phase recognition either suppose that each video is initially segmented into different actions~\cite{SGC} or perform action recognition with low temporal resolution~\cite{DPS}. However, these approaches are incapable of providing automatic frame-wise labeling.
Similarly, some methods assume that particular side information, such as tool presence signals, is available during training~\cite{RTM}. However, relying purely on surgical videos for workflow analysis is preferred, since (1) providing side information by the surgeons will impose burdens on their constraint schedule, and (2) RFID data or tool usage signals are not ubiquitous in regular hospitals.    

\vspace{0.50em}
\noindent \textbf(III) The existing RNN architectures such as DeepPhase~\cite{DPS}  are not end-to-end approaches and this may result in suboptimal performance. An end-to-end approach can enable correlated spatio-temporal feature learning between recurrent and spatial convolutional layers. DeepPhase also exploits the information from the previous states to infer the corresponding phase to the current state. Due to the high computational complexity of recurrent neural networks, however, the recurrent layer can be unrolled over a limited number of frames. Since some phases may span for several seconds, in the mentioned schemes, the input frames are sampled with a low frame-rate per second (3fps in DeepPhase). Consequently, these schemes are unable to infer the phases with high temporal resolution.

\vspace{0.50em}
\noindent \textbf(IV) To perform classification on barely separable data, more complicated features, and therefore deeper neural network architectures are required. However, a deep neural network entails more annotations and is more vulnerable to overfitting. To obtain higher accuracy with fewer image annotations, the networks are trained on low-resolution inputs. A serious defect of these approaches is the distortion of the relevant content during image downsampling. Regarding phase recognition in cataract surgery videos, these relevant contents include the instruments and cornea. The distortion inflicted by downsampling can negatively affect the classification performance.

\vspace{0.50em}
\noindent \textbf(V) Finally, due to class imbalance in some datasets~\cite{DPS}, the classification results cannot accurately reflect the performance of the trained networks.

\subsection{Mask R-CNN}
Mask Region-based CNN~\cite{MRCNN} (Mask R-CNN) is designed to perform instance segmentation by dividing it into two sub-problems: (1) object detection that is the process of localizing and classifying each object of interest, and (2) semantic segmentation that handles the delineation of the detected object at pixel-level. In the object detection module, a region proposal network (RPN) uses the low-resolution convolutional feature map (CFM) coming from a backbone network. RPN attaches nine different anchors centered around each feature vector in CFM. These anchors have three different aspect ratios to deal with different object types (horizontal, vertical, or squared) and three different scales to deal with scale variance (small, medium, and large-sized objects). The anchor properties along with the computed features corresponding to each anchor are used to decide the most fitted bounding box for each object of interest. In the semantic segmentation module, a fully convolutional network (FCN) is utilized to output instance segmentation results for each object of interest. Using the low-resolution detection of the object detection module, the FCN produces the masks with the same resolution as the original input of the network. 

\begin{figure*}[!th]
    \centering
    \includegraphics[width=1\textwidth]{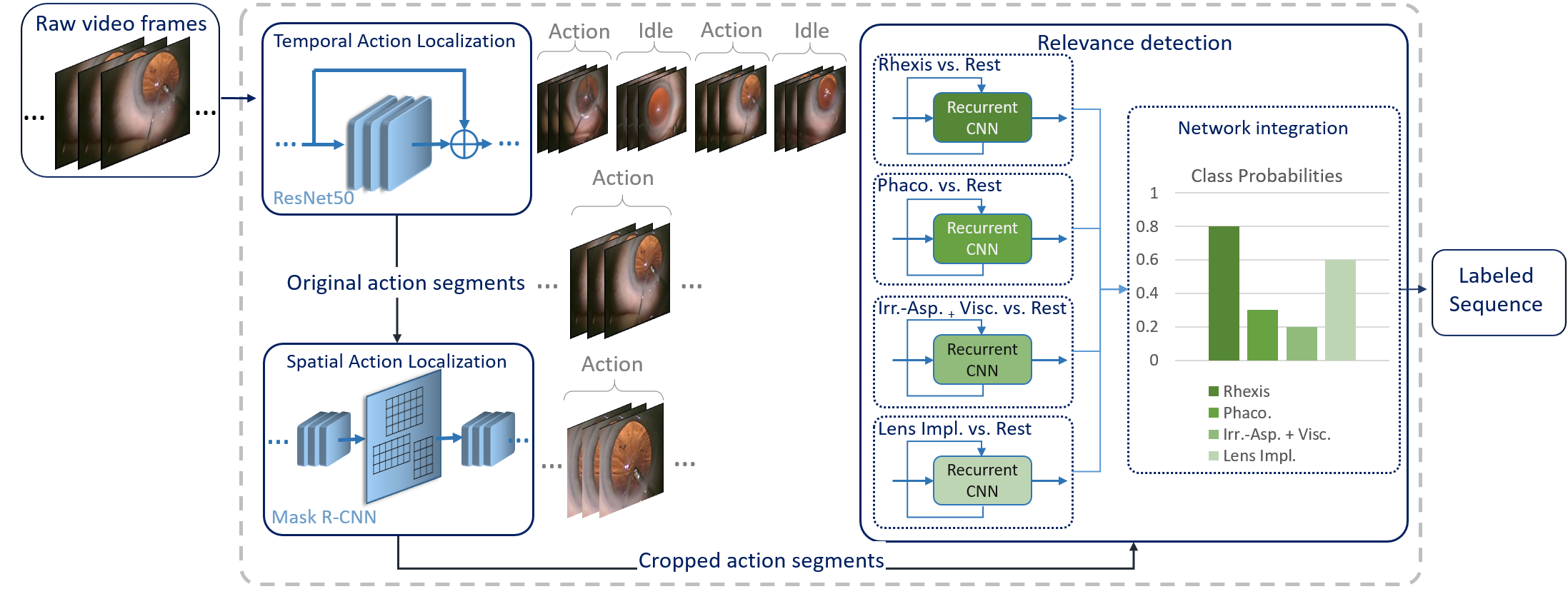}
    \caption{Block diagram of the proposed approach.}
    \label{fig: Block_diagram}
\end{figure*}

\subsection{Proposed Approach}
Fig.~\ref{fig: Block_diagram} demonstrates the overview of the proposed framework which consists of three modules:

\vspace{0.5\baselineskip}
\noindent\textit{Temporal action localization: }
%To reduce the false-negative rate arising from motion and defocus blur, we further use a temporal averaging filter with 
 Each action phase in cataract surgery is always delimited by two idle phases. An \textit{idle phase} refers to a temporal segment in a cataract surgery video where no instrument is visible inside the frame -- and accordingly, no surgical action is performed. Detecting the idle phases can enhance relevance detection results by enabling the use of complementary spatio-temporal information from the same action phase. We propose to use a static residual network to categorize the frames of cataract surgery into \textit{action} or \textit{idle} frames. This pre-processing step plays a crucial role in alleviating phase classification in the \textit{relevance detection} module. 

\vspace{0.5\baselineskip}
\noindent\textit{Spatial action localization:}
The rationale behind spatial action localization is to mitigate the effect of the low-resolution input image on classification performance while retaining all discriminative and informative content for training. Since all the relevant phases in cataract surgery occur inside the cornea, higher resolution of the cornea can significantly boost the classification results. One way to provide a higher-resolution cornea for a network with a particular input size is to detect the cornea, crop the bounding box of the cornea, and use the cropped version instead of the original frame as the input of the network. In addition to providing high-resolution relevant content, this localization approach results in eliminating the redundant information that can cause network overfitting during training. We suggest using \textit{Mask R-CNN} as the state-of-the-art approach in instance-segmentation to detect the cornea and then resize the bounding box of the cornea to fit the input size of the relevance detection module.
%~\footnote{It should be noted that the bounding boxes extracted based on mask segmentation are different from the output bounding boxes of object detection networks and much more accurate.}.
%Indeed, the network can take more advantage of the high-resolution relevant spatial content while being prevented from learning irrelevant spatial content.

\vspace{0.5\baselineskip}
\noindent\textit{Relevance Detection: }
For this module, we propose a recurrent CNN to be trained on the spatiotemporally localized action segments of cataract videos. The network is provided with sequences of the relevance-based cropped frames that contain higher-resolution information for the cornea region and complementary temporal information from the neighboring frames. In an end-to-end training manner, the network can benefit from the seamless integration of spatio-temporal features. Moreover, using back-propagation through time, the recurrent layer encourages the spatial descriptors to learn the shared representations among the input frames while preventing them from learning exceptional irrelevant features. We have experimentally found out that integrated \textit{one-vs-rest} networks provide higher accuracy compared to multi-class classification networks. Thus we propose to train four parallel recurrent networks (Fig.~\ref{fig: Block_diagram}), each one being responsible for detecting one particular relevant phase.  Fig~\ref{fig: Recurrent_model} demonstrates the configuration of the proposed network for relevance detection. The network contains a time distributed residual CNN (ResNet101 in our experiments) that outputs a sequence of spatial feature maps for the sequence of input frames. The network is trained one time per each relevant phase (to detect the relevant phase versus rest). The results of the four networks trained to detect four relevant phases are further integrated by using their output class probabilities in addition to the classification results. If an input is classified as the relevant phase in more than one network, the relevant phase with the highest probability is chosen as the corresponding class to that input.

\section{Experimental Settings}
\label{sec: Experimental Setup}
\subsection{Alternative approaches}
For relevance detection, we have implemented and compared many approaches that can be categorized into (i) \textit{static convolutional networks}, (ii) \textit{feature-based CNN-RNN} -- here, we first train the CNN independently, then replace the output layer with recurrent layers (all the layers of CNN are frozen during training the RNN layers), and (iii) \textit{end-to-end CNN-RNN} -- in contrast with feature-based CNN-RNN, the CNN is not frozen in end-to-end training manner. 

To provide a fair comparison, all the networks are trained and tested on the same dataset being created based on the results of the \textit{temporal action localization} module. In the proposed approach, however, we further pass the dataset through the \textit{spatial action localization} module to prove the effectiveness of this module in enhancing the model accuracy.

\vspace{0.5\baselineskip}
\subsection{Dataset}
%The publicly available Cataract-101 dataset~\cite{CAT101},  collected in 2017-2018 at \textit{Klinikum Klagenfurt} (Austria), is used as the basis for our experiments. 

Together with clinicians from \textit{Klinikum Klagenfurt} (Austria), we recorded videos from 22 cataract surgeries and annotated medically relevant phases. The dataset (with all videos and annotations) is publicly released with this work in the following link: \href{https://ftp.itec.aau.at/datasets/ovid/relevant-cat-actions/}{https://ftp.itec.aau.at/datasets/ovid/relevant-cat-actions/}.

\vspace{0.5\baselineskip}
\noindent{\textit{Temporal action localization: }}
For this step, all frames of 22 videos from the dataset are annotated and categorized as \textit{idle} or \textit{action} frames. We have used 18 randomly selected videos from the annotations for training and the remaining videos are used for testing. To prepare a balanced dataset for both training and testing stages, 500 idle and 500 action frames are uniformly sampled from each video, composing 9000 frames per class in the training set and 2000 frames per class in the testing set.

\vspace{0.5\baselineskip}
\noindent{\textit{Spatial action localization: }}
The area of the \textit{cornea} in 262 frames from 11 cataract surgery videos is annotated for the cornea detection network. The network is trained using 90\% of the annotations and tested on the remaining 10\%.

\vspace{0.5\baselineskip}
\noindent{\textit{Relevance detection: }}
For this module, all the action segments of 10 videos are annotated and categorized as \textit{rhexis}, \textit{phacoemulsification} (termed as \textit{Phaco.}), \textit{irrigation-aspiration} with \textit{viscoelastic-suction} (\textit{Irr.-Asp.+Visc.}), \textit{lens implantation} (\textit{Lens Impl.}), and the remaining content (\textit{Rest}). From these annotations, eight videos are randomly selected for training and two other videos are used for testing. Since recurrent CNNs require a sequence of images as input, we have created a video dataset using the annotated segments. Each annotated segment is decoded and 75 successive frames (three seconds) with a particular overlapping step are losslessly encoded as one input video. Due to different average duration of different relevant phases, we use a different overlapping step for short relevant phases to yield a balanced dataset. This overlapping step is one frame for the rhexis and implantation phase, and four frames for other phases. Afterward for each network, 2000 clips per class from the training videos and 400 clips per class from the test videos are uniformly sampled. This amounts to 4000 clips from eight videos as the training set and 800 clips from two other videos as the testing set.

\subsection{Neural Network Models} 
\noindent\textit{Temporal action localization: }For idle-frame recognition, we have exploited ResNet50 and ResNet101~\cite{ResNet50} pre-trained on ImageNet~\cite{imageNet}. Excluding the top of these networks, the average pooling layer is followed by a \textit{dropout} layer with its dropping probability being equal to 0.5. Next, a \textit{dense} layer with two output neurons and \textit{softmax} activation is added to the network to form the output layer. The classification performances of these networks are compared and the network with the best performance is used for later experiments.

\vspace{0.5\baselineskip}
\noindent\textit{Spatial action localization: }For cornea detection, we utilize the \textit{Mask R-CNN} network~\cite{MRCNN, matterport_maskrcnn_2017}. We train the network on two different backbones (ResNet50 and ResNet101) and use the backbone with the best results to produce the cropped input for the relevance detection module.

\begin{figure}[!t]
    \centering
    \includegraphics[width=0.85\columnwidth]{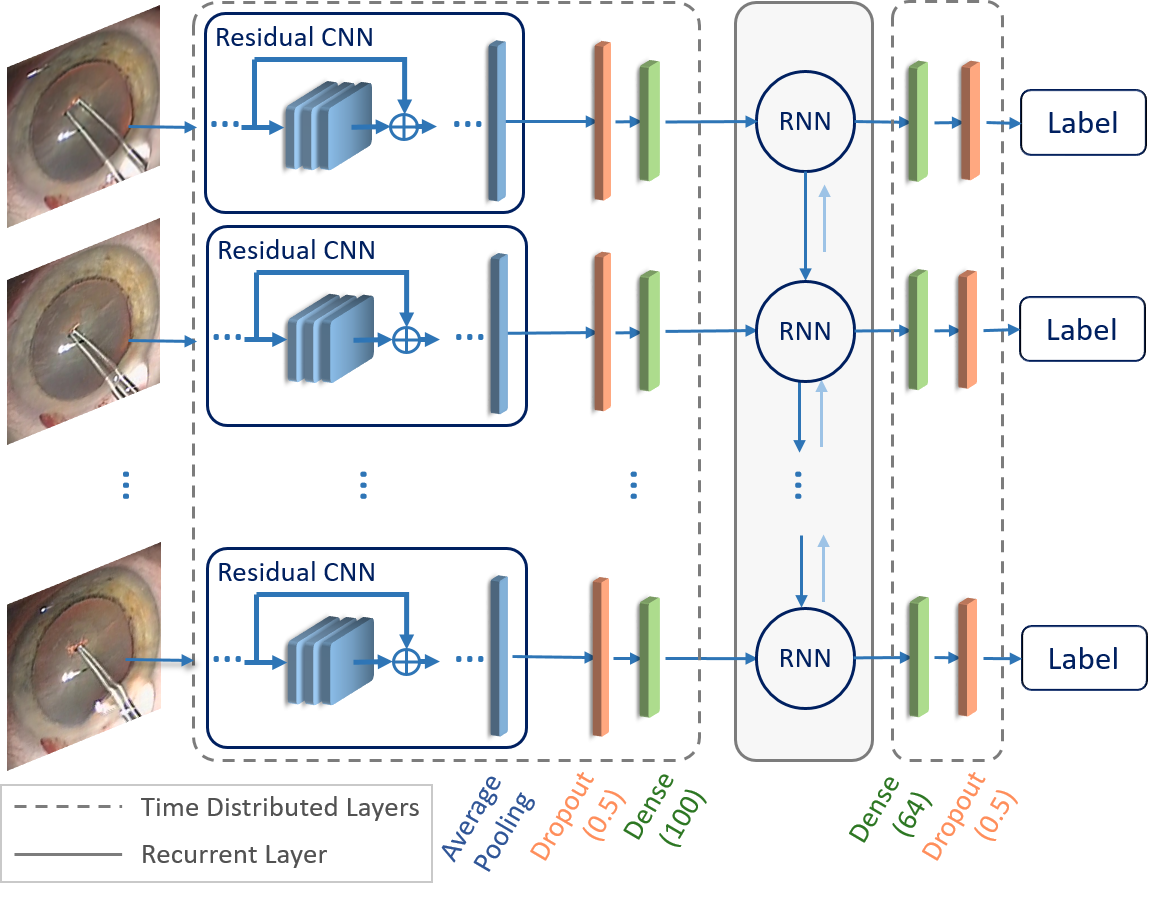}
    \caption{Schematic of the proposed CNN-RNNs for relevance detection.}
    \label{fig: Recurrent_model}
    
\end{figure}

\vspace{0.5\baselineskip}
\noindent\textit{Relevance detection: }For static CNNs, we have used ResNet50 and ResNet101 with the same configuration as for the \textit{temporal action localization} module. The network with the best results is then used as the baseline for both feature-based and end-to-end recurrent networks. Fig.~\ref{fig: Recurrent_model} shows the shared schematics of CNN-RNNs. In feature-based models, the average-pooling layers of the trained static models are used as a backbone by freezing all of the CNN layers. In contrary to feature-based models, we train the static and recurrent layers of the end-to-end models simultaneously and by starting from the weights initialized from ImageNet. We have compared four different recurrent models: (1) CNN+LSTM in which the recurrent layer includes one LSTM layer, (2) CNN+GRU which contains a GRU layer, (3) CNN+BiLSTM that utilizes a bidirectional LSTM layer, and (4) CNN+BiGRU containing a bidirectional GRU layer. All of the recurrent layers contain five units.

\subsection{Neural Network Settings}
\noindent\textit{Temporal action localization: }The SGD optimizer with $decay = 1e-6$ and $momentum = 0.9$ is set as the optimization function during training. The temporal action localization network is trained for 10 epochs with the initial learning rate $lr_1$ being set to $0.0005$.  The network is then fine-tuned for 10 epochs with $lr_2 = lr_1/5$, and 10 other epochs with $lr_3 = lr_1/10$. To avoid network overfitting, all the layers except for the last 20 layers are frozen during training. Also, \textit{categorical cross-entropy} is used as the loss function.

\vspace{0.5\baselineskip}
\noindent\textit{Spatial action localization: } 
The \textit{Mask R-CNN} network pre-trained on the COCO dataset~\cite{COCO} is fine-tuned in an end-to-end manner starting with learning rate being equal to $0.001$. The network is trained for 50 epochs; the initial learning rate is divided by 2, 10, 20 and 100 after epochs 10, 20, 30, and 40, respectively.  

\vspace{0.5\baselineskip}
\noindent\textit{Relevance detection: } 
Table~\ref{Tab: hyperparameters} details the settings of hyper-parameters for the proposed relevance detection approach and the rival neural networks simulated to evaluate the effectiveness of the proposed approach. We have performed extensive hyperparameter optimizations to achieve a feasible speed in training while preventing overfitting during training. We came up with different numbers of frozen layers and different learning rates. For instance, the initial learning rate for static networks is set to $10^{-4}$. This learning rate is divided by 2, 10, and 20 after 2, 20, and 30 epochs respectively. Besides, we use the same settings for the SGD optimizer in this module as for the \textit{temporal action localization} module. 
\begin{table}[th!]
\renewcommand{\arraystretch}{1.1}
\caption{Data augmentation methods applied to the classification and segmentation networks.}
\label{tab:aug}
\centering
\begin{tabu}{lcc}
\specialrule{.12em}{.05em}{.05em}%\tabucline[1pt]{1-4}
Augmentation Method&Property&Value\\\specialrule{.12em}{.05em}{.05em}%\hline
Brightness&Value range&[-50,50]\\
Gamma contrast&Gamma coefficient&[0.5,2]\\
Gaussian blur&Sigma&[0.0, 5.0]\\
Motion blur&Kernel size&9\\
Crop and pad&Percentage&[-0.25,0.25]\\
Affine&Scaling percentage&[0.5,1.5]\\
\specialrule{.12em}{.05em}{0.05em}%\tabucline[1pt]{1-4}
\end{tabu}
\end{table}
\begin{table}[thpb!]
\renewcommand{\arraystretch}{1}
\caption{Precision, Recall, and F1-Score of \textit{temporal action localization} module.}
\label{Tab: Classification_report_Idle}
\centering
\begin{tabularx}{0.9\columnwidth}{l c c c c c}
\specialrule{.12em}{.05em}{.05em}%\tabucline[1pt]{1-9}
 Network & Class && Precision & Recall & F1-Score \\\specialrule{.12em}{.05em}{.05em}%\tabucline[1pt]{1-9}
 \multirow{3}{*}{ResNet50} & Action& & 1.00 & 0.85 & 0.92 \\
 & Idle && 0.87 & 1.00 & 0.93 \\\cdashline{3-6}[0.6pt/1pt]
 & Macro avg &&  0.93 & 0.92 & 0.92 \\\hline
 \multirow{3}{*}{ResNet101} & Action && 0.99 & 0.88 & 0.93 \\
 & Idle && 0.89 & 0.99 & 0.94 \\\cdashline{3-6}[0.6pt/1pt]
 & Macro avg &&  0.94 & 0.94 & 0.94\\
\specialrule{.12em}{.05em}{.05em}%\tabucline[1pt]{1-9}
\end{tabularx}
\end{table}
Due to the high computational complexity of the end-to-end training approaches, the RNN layer should be just unrolled over a short segment (clip) instead of a complete video. In this study, we assume that the RNN layers can access up to five frames for decision. The sequence generator divides each input video into five temporal segments and randomly chooses one frame from each temporal segment. This random choosing of the frames encourages the network to learn the relationships in diverse temporal distances. It can also act as a temporal data augmentation technique.

\subsection{Data Augmentation Methods}
Data augmentation during training plays a vital role in preventing network overfitting as well as boosting the network performance in case of unseen data. Accordingly, the input frames to all networks are augmented during training using offline and online transformations. Table~\ref{tab:aug} lists the detailed descriptions of augmentation methods utilized for all networks. The listed transformations are selected based on either inherent or statistical features in the dataset. For instance, motion blur and Gaussian blur augmentations are chosen due to having harsh motion blur and defocus blur in our dataset. 

\subsection{Evaluation Metrics}
We report the performances of \textit{temporal action localization} and \textit{relevance detection} networks using the common classification metrics namely precision, recall, accuracy, and F1-score. For the \textit{spatial action localization} network, we evaluate the performance using average precision over recall values with different thresholds for Intersection-over-Union (IoU), as well as mean average precision (mAP) over IoU in the range of 0.5 to 0.95.
\begin{table}[!thp]
\renewcommand{\arraystretch}{1}
\caption{Instance detection and segmentation results of \textit{spatial action localization} module.}
\label{Tab: instance-segmentation}
\centering
\begin{tabularx}{0.4\textwidth}{ l l *{6}{Y} }
\specialrule{.12em}{.05em}{.05em} 
%\toprule
 Backbone &&\multicolumn{3}{c}{Mask Segmentation}\\\specialrule{.12em}{.05em}{.05em} %\midrule%\tabucline[1pt]{1-8}
&&$mAP_{80}$&$mAP_{85}$&$mAP$\\\cdashline{2-8}[0.6pt/1pt]
ResNet101&&1.00&0.92&0.89\\
ResNet50&&1.00&1.00&0.88\\
\specialrule{.12em}{.05em}{.05em} %\bottomrule
 &&\multicolumn{3}{c}{Bounding-Box Segmentation}\\\specialrule{.12em}{.05em}{.05em}
&&$mAP_{80}$&$mAP_{85}$&$mAP$\\\cdashline{2-8}[0.6pt/1pt]
ResNet101&&1.00&1.00&0.95\\
ResNet50&&1.00&1.00&0.94\\
\specialrule{.12em}{.05em}{.05em} %\bottomrule
\end{tabularx}
\end{table}
\begin{figure}[!thpb]
    \centering
    \includegraphics[width=0.8\columnwidth]{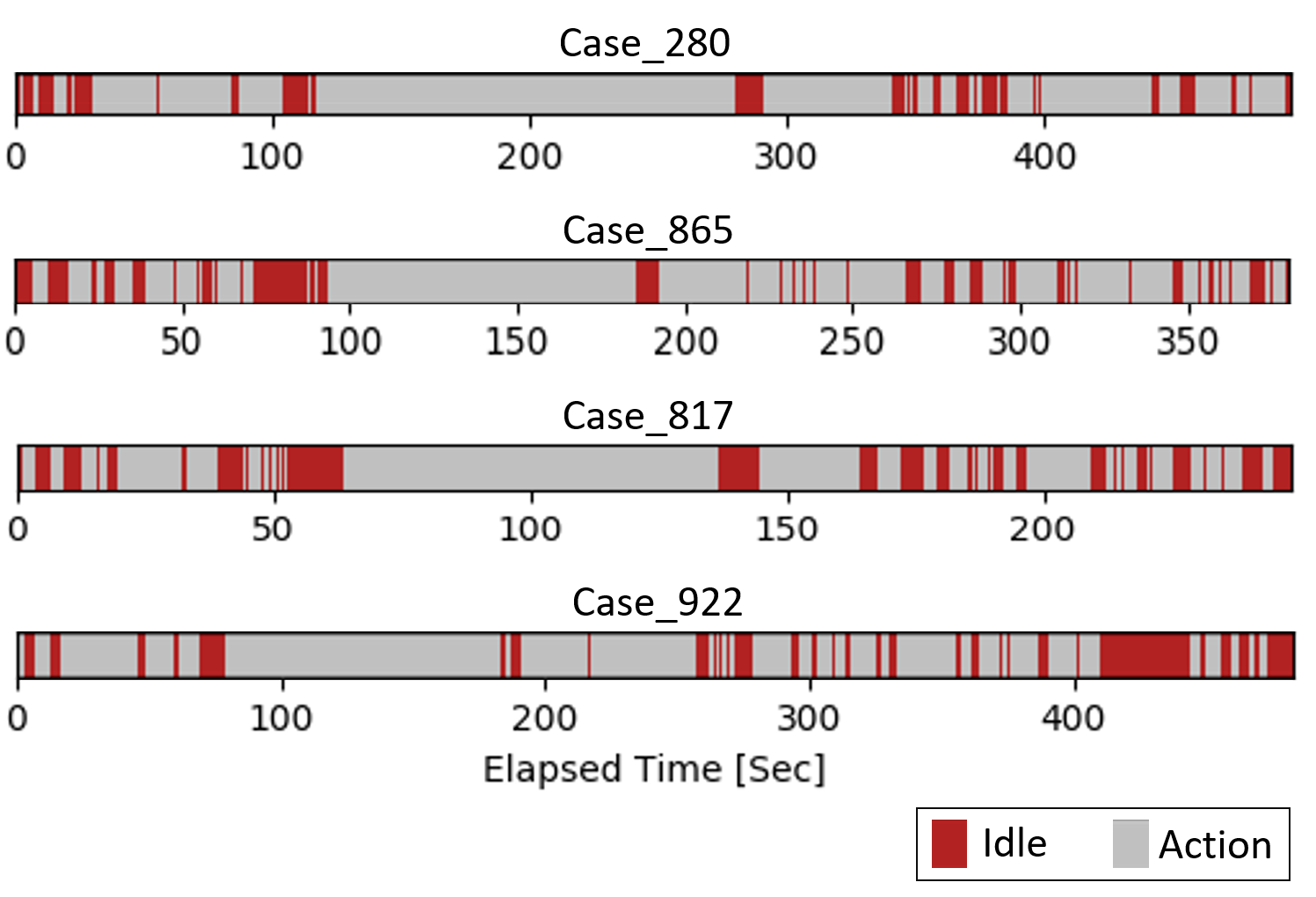}
    \caption{Pattern of \textit{temporal action localization} for four representative videos.}
    \label{fig: Idle-pattern}
\end{figure}

\section{Experimental Results and discussion}
\label{sec: Experimental Results}
\subsection{Temporal Action Localization}
Table~\ref{Tab: Classification_report_Idle} reports the main classification metrics for \textit{temporal action localization} using ResNet50 and ResNet101. As can be perceived, both models are fairly accurate and show a close performance, with the F1-score of ResNet101 being $1\%$ better than ResNet50. Thus we use ResNet101 for further experiments. The reason why we have a lower rate of recall for action phases is rooted in the inherent problems in the dataset. The harsh motion blur in some action frames distorts the instruments' spatial content and makes them even invisible for the human eye. To retrieve these wrong predictions, we use a temporal mean filter with a window size of 15 (around 0.5 seconds) as a post-processing step. Fig.~\ref{fig: Idle-pattern} illustrates the filtered \textit{temporal action localization} results for four representative cataract surgery videos.

\subsection{Spatial Action Localization}
The mask segmentation and bounding-box detection results for cornea tracking are presented in Table~\ref{Tab: instance-segmentation}. It should be noted that the bounding-box segmentation results based on instance segmentation networks are much more accurate compared to that of the object detectors. This is the reason why we use \textit{Mask R-CNN}, although we just need the bounding-box of the cornea. The figures for bounding-box segmentation affirm that both networks can detect the cornea with at least $0.85\%$ IoU. Since the network trained with the ResNet101 backbone shows $1\%$ higher $mAP$ compared to that with ResNet50 backbone, this trained network will be used for further experiments.

\subsection{Relevance Detection}

\begin{table*}[thpb!]
\renewcommand{\arraystretch}{1}
\caption{Training hyperparameters and specification of the proposed and alternative \textit{relevance detection} approaches.}
\label{Tab: hyperparameters}
\centering
\begin{tabularx}{1\textwidth}{ l l l *{8}{c} }
\specialrule{.12em}{.05em}{.05em} 
%\toprule
\multirow{2}{*}{Model} & \multirow{2}{*}{specification} & \multirow{2}{*}{Name} &\multicolumn{5}{c}{Hyper parameters}\\\cdashline{4-8}[0.8pt/1pt] %\midrule%\tabucline[1pt]{1-8}
&&&Frozen-Layers&Optimizer&Epochs&Learning-Rate&Batch-Size\\\specialrule{.12em}{.05em}{.05em}
\multirow{3}{*}{Static}&ResNet50&CNN50&[1:-20]&&&&\\

&ResNet101&CNN101&[1:-10]&SGD&40&$lr = 10^{-4}, \frac{lr}{2}\mid_{2}, \frac{lr}{10}\mid_{20}, \frac{lr}{20}\mid_{30}$&16\\

&ResNet152&CNN152&[1:-10]&&&&\\\hline

\multirow{4}{*}{\begin{tabular}{@{}l@{}}Recurrent\\(feature-based)\\ResNet50 backbone\end{tabular}}&GRU&GRU-FB&\multirow{4}{*}{CNN}&\multirow{4}{*}{SGD}&\multirow{4}{*}{15}&\multirow{4}{*}{$lr = 10^{-4}, \frac{lr}{2}\mid_{5}$}&\multirow{4}{*}{16}\\

&LSTM&LSTM-FB&&&&&\\

&Bidirectional GRU&BiGRU-FB&&&&&\\

&Bidirectional LSTM&BiLSTM-FB&&&&&\\\hline

\multirow{4}{*}{\begin{tabular}{@{}l@{}}Recurrent\\(end to end)\\ResNet50 backbone\end{tabular}}&GRU&GRU-E2E&\multirow{4}{*}{[1:-10]}&\multirow{4}{*}{SGD}&\multirow{4}{*}{15}&\multirow{4}{*}{$lr = 10^{-4}$}&\multirow{4}{*}{16}\\
&LSTM&LSTM-E2E&&&&&\\
&Bidirectional GRU&BiGRU-E2E&&&&&\\
&Bidirectional LSTM&BiLSTM-E2E&&&&&\\\hline

\multirow{2}{*}{\begin{tabular}{@{}l@{}}Recurrent\\
(end to end)\end{tabular}}&Bidirectional GRU&LocalPhase-G&\multirow{2}{*}{[1:-10]}&\multirow{2}{*}{SGD}&\multirow{2}{*}{15}&\multirow{2}{*}{$lr = 10^{-4}$}&\multirow{2}{*}{16}\\
&Bidirectional LSTM&LocalPhase-L&&&&&&&&\\
%\tabucline[1pt]{1-8}
\specialrule{.12em}{.05em}{.05em} %\bottomrule
\end{tabularx}
\end{table*}

\begin{table*}[thpb!]
\renewcommand{\arraystretch}{1.1}
\caption{Precision, Recall, and F1-Score of the proposed and alternative \textit{relevance detection} approaches.}
\label{Tab: precision-recall2}
\centering
\begin{tabularx}{1\textwidth}{ l  *{12}{c} }
\specialrule{.12em}{.05em}{.05em} 
\multirow{2}{*}{Network}&\multicolumn{3}{c}{Rhexis}&\multicolumn{3}{c}{Phaco.}&\multicolumn{3}{c}{Lens Impl.}&\multicolumn{3}{c}{Irr.-Asp.+Visc.}\\\cdashline{2-13}[0.8pt/1pt] 
&Precision&Recall&F1-Score&Precision&Recall&F1-Score&Precision&Recall&F1-Score&Precision&Recall&F1-Score\\\hline
CNN50&0.81&0.81&0.81&0.88&0.85&0.85&0.83&0.82&0.82&0.74&0.55&0.45\\
CNN101&0.82&0.73&0.71&0.86&0.83&0.83&0.77&0.58&0.49&0.72&0.72&0.72\\
CNN152&0.95&0.95&0.95&0.76&0.71&0.69&0.81&0.74&0.72&0.69&0.63&0.60\\\hline
GRU-FB&0.99&0.99&0.99&0.93&0.92&0.92&\textbf{0.88}&0.85&0.84&0.80&0.73&0.71\\
LSTM-FB&0.95&0.95&0.95&0.90&0.90&0.90&0.85&0.80&0.79&0.77&0.77&0.77\\
BiGRU-FB&0.98&0.98&0.98&0.93&0.92&0.92&0.87&0.82&0.79&0.78&0.79&0.67\\
BiLSTM-FB&\textbf{1.00}&\textbf{1.00}&\textbf{1.00}&0.92&0.91&0.91&\textbf{0.88}&0.84&0.83&0.79&0.76&0.76\\\hline
GRU-E2E&0.98&0.98&0.98&0.92&0.91&0.91&0.83&0.75&0.74&0.78&0.66&0.63\\
LSTM-E2E&0.95&0.94&0.94&0.83&0.83&0.83&0.84&0.79&0.78&0.69&0.66&0.64\\
BiGRU-E2E&0.99&0.99&0.99&0.92&0.91&0.91&0.84&0.76&0.75&0.80&0.76&0.75\\
BiLSTM-E2E&\textbf{1.00}&\textbf{1.00}&\textbf{1.00}&0.93&0.93&0.93&0.80&0.67&0.63&0.74&0.71&0.70\\\hline
LocalPhase-G&0.99&0.98&0.98&0.95&0.94&0.94&0.86&0.85&0.85&\textbf{0.85}&0.80&0.80\\
LocalPhase-L&0.99&0.99&0.99&\textbf{0.96}&\textbf{0.96}&\textbf{0.96}&0.87&\textbf{0.86}&\textbf{0.86}&0.84&\textbf{0.83}&\textbf{0.83}\\
\specialrule{.12em}{.05em}{.05em} %\bottomrule
\end{tabularx}
\end{table*}

\begin{table}[thpb!]
\renewcommand{\arraystretch}{1.1}
\caption{Accuracy of the proposed and alternative \textit{relevance detection} approaches.}
\label{Tab: accuracy}
\centering
\begin{tabularx}{0.9\columnwidth}{ lcccc }
\specialrule{.12em}{.05em}{.05em} 
Network&Rhexis&Phaco.&Lens Impl.&Irr.-Asp.+Visc.\\\hline
CNN50&0.81&0.85&0.82&0.55\\
CNN101&0.73&0.83&0.58&0.72\\
CNN152&0.95&0.71&0.74&0.63\\\hline
GRU-FB&0.99&0.92&0.85&0.73\\
LSTM-FB&0.95&0.90&0.80&0.77\\
BiGRU-FB&0.98&0.92&0.82&0.69\\
BiLSTM-FB&\textbf{1.00}&0.91&0.84&0.76\\\hline
GRU-E2E&0.98&0.91&0.75&0.66\\
LSTM-E2E&0.94&0.83&0.79&0.66\\
BiGRU-E2E&0.99&0.91&0.76&0.76\\
BiLSTM-E2E&\textbf{1.00}&0.93&0.67&0.71\\\hline
LocalPhase-G&0.98&0.94&0.85&0.81\\
LocalPhase-L&0.99&\textbf{0.96}&\textbf{0.86}&\textbf{0.83}\\
\specialrule{.12em}{.05em}{.05em} %\bottomrule
\end{tabularx}
\end{table}

The classification reports of the different static, feature-based recurrent, and end-to-end recurrent neural networks are listed in Table~\ref{Tab: precision-recall2} and Table~\ref{Tab: accuracy}. Considering the static CNNs (namely ResNet50, ResNet101, and ResNet152), we can see different behaviors of a network for different phases. This difference lies in the level of similarity between each target phase and other phases. For instance, the  \textit{Irr.-Asp.+Visc.} phase shares a lot of statistics and visual similarities with the \textit{Phaco.} phase. Since the frames corresponding to \textit{Phaco.} contain more feature variations, all static CNNs tend to classify \textit{Irr.-Asp.+Visc.} frames as \textit{Phaco.} This tension decreases by increasing the number of layers in the network, as it contributes to discriminating more complicated features. On the other hand, networks with more parameters are more prone to overfit during training on small datasets. In summary, ResNet50 and ResNet101 have shown the same level of accuracy on average. Thus we choose ResNet50 having fewer parameters as the baseline for the recurrent networks.

Thanks to the \textit{temporal action localization} module, the feature-based recurrent neural networks have shown noticeable enhancement in classification results, specifically for \textit{rhexis} and \textit{Phaco.} phase. Interestingly, the bidirectional LSTM network can retrieve 100\% of the frames corresponding to the \textit{rhexis} phase. In summation, it can be observed that all the different configurations of the feature-based recurrent networks outperform the static CNNs. 
%Using recurrent layers that enable joint spatio-temporal feature learning (optimization)
Regarding the end-to-end training approaches, we can notice some drops in the classification results for \textit{Irr.-Asp.+Visc.} phase and \textit{Lens Impl.} phase. This drop can occur due to an insufficient number of training examples. The end-to-end training approaches are more vulnerable to overfitting due to their high degree of freedom. 
%can improve the consistency of the networks' output and boost the classification performance.

Both configurations of the proposed approach (namely bidirectional GRU and bidirectional LSTM) have achieved superior performance compared to the alternative approaches. Also, it can be perceived from the Table~\ref{Tab: accuracy} that our models have the best accuracy in detecting the \textit{Irr.-Asp.+Visc.} that is the most challenging phase to retrieve. With completely identical configuration to BiGRU-E2E, LocalPhase-G which takes advantage of the \textit{spatial action localization} module has achieved more reliable results (3\% gain in F1-score for \textit{Phaco.} phase and 5\% percent gain in F1-score for \textit{Irr.-Asp.+Visc.} phase). Likewise, LocalPhase-L has achieved 3\%  and 13\% higher F1-score for the \textit{Phaco.} and \textit{Irr.-Asp.+Visc.} phase, respectively. These results reveal the influence of high-resolution relevant content on network training as well as the effect of redundant-information elimination on preventing network overfitting.

%It should be noted that we cannot directly compare our approach to existing work, since we are the first to (1) focus on medically relevant phases only, and (2) use a completely balanced dataset. When comparing to the most similar approach \cite{DPS}, our results are significantly higher though -- the proposed approach (LocalPhase-L) achieves an average F1-score of 90.5\% (vs. 74.92\%\cite{DPS}) and an  average Accuracy of 91\% (vs. 78.28\%\cite{DPS}), as shown in  Table~\ref{Tab: precision-recall2} and Table~\ref{Tab: accuracy}. 

\section{Conclusion}
\label{sec: Conclusion}
Today, considerable attention from the computer-assisted intervention community (CAI) is focused on enhancing the surgical outcomes and diminishing the potential clinical risks through context-aware assistant or training systems. The primary requirement of such a system is a surgical phase segmentation and recognition tool. In this paper, we have proposed a novel framework for relevance detection in cataract surgery videos to address the shortcomings of the existing phase recognition approaches. Indeed, the proposed approach is designed to (i) work independently of any metadata or side information, (ii) provide relevance detection with a high temporal resolution, (iii) be able to detect the relevant phases notwithstanding the irregularities in the order or duration of the phases, and (iv) be less prone to overfitting in case of the small non-diverse training sets. To alleviate the network convergence and avoid network overfit on small training sets, we have proposed to localize the spatio-temporal segments of each action phase. A recurrent CNN is then utilized to take advantage of this complementary spatio-temporal information by simultaneous training of static and recurrent layers. Experimental results confirm that the networks trained on the relevant spatial regions are more robust against overfitting due to substantially less misleading content. Besides, we have presented the first systematic analysis of recurrent CNN frameworks for phase recognition in cataract surgeries that further confirms the superiority of the proposed approach. 

% Be able INSTEAD OF being able
% Be less INSTEAD OF being less

% eliminate the redundant information of the input frames causing network over-fitting
% To systematically advance the process of surgical training,
% A surgical video exploration system built on machine-learning-assisted video segmentation and indexing

\section*{Acknowledgment}
This work was funded by the FWF Austrian Science Fund under grant P 31486-N31.

%\newpage
\balance
\bibliographystyle{ieeetr}
\bibliography{bibtex.bib}

\end{document}